\documentclass[11pt,a4paper]{article}
\usepackage[hyperref]{acl2021}
\usepackage{times}
\usepackage{latexsym}

\usepackage{microtype}

\aclfinalcopy %

\usepackage{graphicx}
\usepackage{amsmath,amssymb} 
\usepackage{booktabs}
\usepackage{tabularx}
\usepackage{subfig}
\usepackage{enumitem}
\usepackage{siunitx}
\usepackage{amsmath}
\newcommand*{\numOne}[1] {\num[round-precision=2]{#1}}

\definecolor{fancyorange}{RGB}{253, 150, 68}
\definecolor{fancypurple}{RGB}{136, 84, 208}

\newcommand{\rev}[1]{{\color{black}{#1}}}

\title{Enriched Pre-trained Transformers\\ for Joint Slot Filling and Intent Detection}

\author{
     Momchil Hardalov$^1$ \quad Ivan Koychev$^1$ \quad Preslav Nakov$^2$ \\
   $^1$Sofia University ``St. Kliment Ohridski'', Bulgaria, \\
   $^2$Qatar Computing Research Institute, HBKU, Qatar, \\
   {\tt \{hardalov, koychev\}@fmi.uni-sofia.bg} \\
   {\tt pnakov@hbku.edu.qa}
}

\date{}

\begin{document}
\maketitle

\begin{abstract}
Detecting the user's intent and finding the corresponding slots among the utterance's words are important tasks in natural language understanding. Their interconnected nature makes their joint modeling a standard part of training such models. Moreover, data scarceness and specialized vocabularies pose additional challenges. Recently, the advances in pre-trained language models, namely contextualized models such as ELMo and BERT have revolutionized the field by tapping the potential of training very large models with just a few steps of fine-tuning on a task-specific dataset. Here, we leverage such models, namely BERT \rev{and RoBERTa}, and we design a novel architecture on top of \rev{them}. Moreover, we propose an intent pooling attention mechanism, and we reinforce the slot filling task by fusing intent distributions, word features, and token representations. The experimental results on standard datasets show that our model outperforms both the current non-BERT state of the art as well as some stronger BERT-based baselines. 

\end{abstract}
\section{Introduction}

With the proliferation of portable devices, smart speakers, and the evolution of personal assistants, such as Amazon's Alexa, Apple's Siri, Google's Assistant, and Microsoft's Cortana, a need for better natural language understanding (NLU) has emerged. The major challenges such systems face are \textit{(i)}~finding the intention behind the user's request, and~\textit{(ii)}~gathering the needed information to complete it via slot filling, while \textit{(iii)}~engaging in a dialogue with the user. Table~\ref{tab:task:example} shows a user request collected from a personal voice assistant. Here, the intent is to \textit{play music} by the artist \textit{Justin Broadrick} from year \textit{2005}.
The slot filling task naturally arises as a sequence tagging task. Conventional neural network architectures, such as RNNs or CNNs are appealing approaches to tackle the problem. 

\begin{table}[t!]
    \centering
    \footnotesize
    \setlength{\tabcolsep}{2.65pt}
    \begin{tabularx}{\linewidth}{lccccccc}
        \toprule
        \bf{Intent} & \multicolumn{6}{c}{PlayMusic} \\
        \midrule
        \bf{Words} & play & music & from & \textcolor{fancyorange}{2005} & by & \textcolor{fancypurple}{justin} & \textcolor{fancypurple}{broadrick}  \\
        & $\downarrow$ & $\downarrow$ & $\downarrow$ & $\downarrow$ & $\downarrow$ & $\downarrow$ & $\downarrow$ \\
        \bf{ Slots} & O & O & O & \textcolor{fancyorange}{B-year} & O & \textcolor{fancypurple}{B-artist} & \textcolor{fancypurple}{I-artist} \\
        \bottomrule 
    \end{tabularx}
    \caption{Example from the SNIPS dataset with slots encoded in the BIO format. The utterance's intent is \textit{PlayMusic}, and the given slots are \textit{year} and \textit{artist}.}
    \label{tab:task:example}
\end{table}{}

Various extensions thereof can be found in previous work~\citep{xu2013convolutional,goo-etal-2018-slot,hakkani-tr2016multi-domain, liu2016attention,e-etal-2019-novel,gangadharaiah-narayanaswamy-2019-joint}. Finally, sequence tagging approaches such as Maximum Entropy Markov model (MEMM)~\citep{toutanvoa-manning-2000-enriching,mccallum2000maximum} and Conditional Random Fields (CRF)~\citep{Lafferty:2001:CRF:645530.655813,jeong2008triangular,huang2015bidirectional} have been added on top to enforce better modeling of the slot filling task. Recently, some interesting work has been done with hierarchical structured capsule networks~\citep{xia-etal-2018-zero,zhang-etal-2019-joint}. %

Here, we investigate the usefulness of pre-trained models on the task of NLU. Our approach is based on BERT~\citep{devlin-etal-2019-bert} \rev{and its successor RoBERTa~\citep{liu2019roberta}}. That model offer two main advantages over previous work~\citep{hakkani-tr2016multi-domain,xu2013convolutional,gangadharaiah-narayanaswamy-2019-joint,liu2016attention,e-etal-2019-novel,goo-etal-2018-slot}: (\textit{i})~they are based on the Transformer architecture~\citep{NIPS2017_7181:transformer}, which allows them to use bi-directional context when encoding the tokens instead of left-to-right (as in RNNs) or limited windows (as in CNNs), and (\textit{ii}) the model is trained on huge unlabeled text collections,
which allows it to leverage relations learned during pre-training, e.g.,~that \textit{Justin Broadrick} is connected to music or that \textit{San Francisco} is a city.

\begin{figure*}[t!]
    \centering
    {
    \subfloat[BERT-Joint.\label{subfig:base:bert}]{%
        \includegraphics[width=0.35\textwidth]{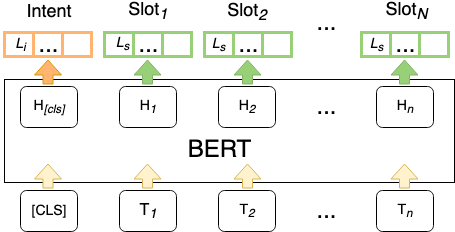}
    }
    \hspace{1.5cm}
    \subfloat[Transformer-NLU (ours). \label{subfig:bert:ours}]{%
       \includegraphics[width=0.5\textwidth]{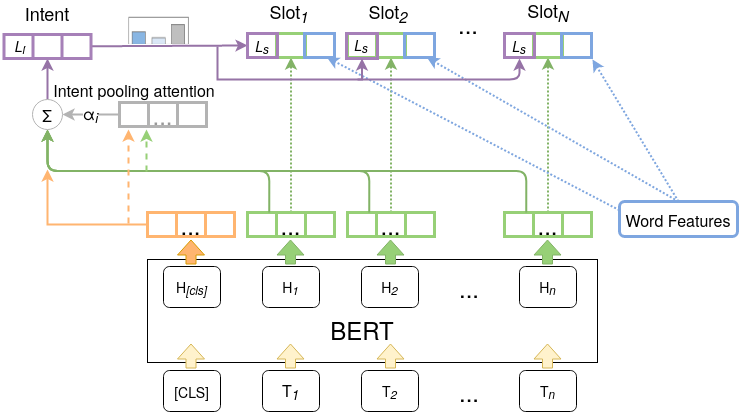}
    }
    }
    \caption{Model architectures for joint learning of intent and slot filling: \subref{subfig:base:bert} classical joint learning with BERT\rev{/RoBERTa}, and
    \subref{subfig:bert:ours} proposed enhanced version of the model.
    }
    \label{fig:models}
\end{figure*}

We further adapt the pre-trained models for the NLU tasks. For the intent, we introduce a pooling attention layer, which uses a weighted sum of the token representations from the last language modelling layer. We further reinforce the slot representation with the predicted intent distribution, and word features such as predicted word casing, and named entities. To demonstrate its effectiveness, we evaluate it on two publicly available datasets: ATIS~\citep{hemphill-etal-1990-atis} and SNIPS~\citep{coucke2018snips}.
Our contributions are as follows:
\begin{itemize}
    \item We enrich pre-trained language model, e.g. BERT and RoBERTa to jointly solve intent classification and slot filling.
    \item We introduce an additional pooling network from the intent classification task, allowing the model to obtain the hidden representation from the entire sequence.
    \item We use the predicted user intent as an explicit guide for the slot fitting layer rather than just depending on the language model
    \item We reinforce the slot learning with features such as named entity, true case annotations.
    \item We present exhaustive analysis of the task-related knowledge, for both datasets, contained in the pre-trained model.
 \end{itemize}

\section{Proposed Approach}
\label{sec:approach}

We propose a joint approach for intent classification and slot filling built on top of a pre-trained language model, i.e.,~BERT~\citep{devlin-etal-2019-bert} \rev{or RoBERTa~\citep{liu2019roberta}}. We further improve the base model in three ways:  (\textit{i})~for intent \rev{detection}, we obtain a pooled representation from the last hidden states for all tokens (Section~\ref{sec:model:intent}),
(\textit{ii})~we obtain predictions for the word case and named entities for each token (word features), and
(\textit{iii})~we feed the predicted intent distribution vector, BERT's last hidden representations, and word features into a slot filling layer (see Section~\ref{sec:model:slot}).
The  %
\rev{architecture} of the model is shown in Figure~\ref{subfig:bert:ours}.

\subsection{Intent Pooling Attention}
\label{sec:model:intent}

Traditionally, BERT and subsequent BERT-style models use a special token ([CLS]) to denote the beginning of a sequence. In the original paper~\citep{devlin-etal-2019-bert}, the authors attach a binary classification loss to it for predicting whether two sequences follow each other in the text (next sentence prediction, or NSP). Adding such an objective forces the last residual block to pool a contextualized representation for the whole sentence from the penultimate layer, which should have a more semantic, rather then task-specific meaning. The latter strives to improve downstream sentence-level classification tasks. However, its effectiveness has been recently debated in the literature~\citep{lample2019cross,joshi2019spanbert,yang2019xlnet,lan2019albert}. It has been even argued that it should be removed \citep{liu2019roberta}.

Here, the task is to jointly learn two strongly correlated tasks, as using representations from the last, task-specific layer is more natural compared to using the pooled one from the [CLS] token. Therefore, we introduce a pooling attention layer to better model the relationship between the task-specific representations for each token and for the intent. We further adopt a global concat attention~\citep{luong-etal-2015-effective} as a throughput mechanism.
Namely, we learn an alignment function to predict the attention weights $\alpha_{int}$ for each token. We obtain the latter by multiplying the outputs from the language model $H \in \mathbb{R}^{N \times d_h}$ by a latent weight matrix~$W_{int\_e} \in \mathbb{R}^{d_h \times d_h}$, where $N$ is the number of tokens in an example and $d_h$ is the hidden size of the Transformer.
This is followed by a non-linear $tanh$ activation. In order to obtain importance logit for each token, we multiply the latter by a projection vector $v \in \mathbb{R}^{d_h}$ (shown in Eq.~\ref{eq:fattn}). We further normalize and scale~\citep{NIPS2017_7181:transformer} to obtain the attention weights. 
\begin{align}
    \label{eq:fattn} 
    align(H) = v \cdot \tanh(W_{int\_e} \cdot H^{T}) \\
    \label{eq:att:intent:alpha} \
    \alpha_{int} = softmax(\frac{align(H)}{\sqrt{d_h}}) \\
    \label{eq:hidden:intent}
    h_{int} = tanh(\sum_{i=1}^{N}{\alpha_{int}^i h_{enc}^i}) \\
    \label{eq:intent:proj}
    y_{int} = W_{int} h_{int}^{T} + b_{int}
\end{align}

Finally, we gather a hidden representation $h_{int}$ as a weighted sum of all attention inputs, and we pass it through a $tanh$ activation (see Eq.~\ref{eq:hidden:intent}). For the final prediction, we use a linear projection on top of $h_{int}$. Finally, we apply dropouts on $h_{int}$, and on the attention weights~\cite{NIPS2017_7181:transformer}.

\subsection{Slots Modeling}
\label{sec:model:slot}

The task of slot filling is closely related to tasks such as part-of-speech (POS) tagging and named entity recognition (NER).
Also, it can benefit from knowing the interesting entities in the text. Therefore, we reinforce the slot filling with tags found by a named entity recognizer. 
Next, we combine the intent prediction, the language model's hidden representations, and some extracted word features into a single vector used for token slot attribution. Details about all components are discussed below.

\paragraph{Word Features} 
\label{sec:model:features}

A major shortcoming of having free-form text as an input is that it tends not to follow basic grammatical principles or style rules. The casing of words can also guide the models while filling the slots, i.e.,~upper-case words can refer to names or to abbreviations. Also, knowing the proper casing enabled the use of external NERs or other tools that depend on the text quality.

As a first step, we improve the text casing using a \textit{TrueCase}
model. The model maps the words into the following classes: \textit{UPPER, LOWER, INIT\_UPPER, and O}, where \textit{O} is for mixed-case words such as \emph{McVey}. With the text re-cased, we further extract the named entities with a NER annotator. Named entities are recognized using a combination of 
three CRF sequence taggers trained on various corpora. 
Numerical entities are recognized using a rule-based system. Both \rev{models}
are part of the 
CoreNLP toolkit~\citep{manning-EtAl:2014:P14-5}. 

Finally, we merge some entities ((job) title, ideology, criminal charge) into a special category \textit{other} as they do not correlate directly to the domains of either dataset. Moreover, we add a custom regex-matching entry for \textit{airport\_code}, which are three-letter abbreviations of the airports. The latter is specially designed for the ATIS~\citep{tur2010left} dataset. While, marking the proper terms, some of the codes introduce noise, e.g.,~the proposition \textit{for} could be marked as an \textit{airport\_code} because of \textit{FOR (Aeroporto Internacional Pinto Martins,
Brazil)}. In order to mitigate this effect, we do a lookup in a dictionary of English words, and if a match is found, we trigger the \textit{O} class for the token. 

In order to allow the network to learn better feature representations for the named entities and the casing, we pass them through a two-layer feed-forward network. The first layer is shown in Eq.~\ref{eq:words:prelu:1} followed by a non-linear PReLU activation, 
where $W_w  \in \mathbb{R}^{23 \times 32}$. The second one is a linear projection $f_{words}$ (Eq.~\ref{eq:words:proj}), where $W_{proj} \in \mathbb{R}^{32 \times 32}$.
\begin{align}
\small
    \label{eq:words:prelu:1} h_{w}^i = PReLU(W_{w} [ners;cases] + b_{w}) \\
    \label{eq:words:proj} f_{words}(ners, cases) =  W_{proj} {h_{w}^i}^{T}  + b_{proj}
\end{align}{}
\vspace{-0.2in}
\paragraph{Sub-word Alignment}
Modern NLP approaches suggest the use of sub-word units~\citep{sennrich-etal-2016-neural,wu2016google,kudo-richardson-2018-sentencepiece}, which mitigate the effects of rare words, while preserving the efficiency of a full-word model. Although they are a flexible framework for tokenization, sub-word units require additional bookkeeping for the models in order to maintain the original alignment between words and their labels.

We first split the sentences into the original word-tag pairs, we then disassemble each one into word pieces \rev{(or BPE, in the case of RoBERTa)}. Next, the original slot tag is assigned to the first word piece, while each subsequent one is marked with a special tag (\textit{X}). Still, the word features from the original token are copied to each unit. To align the predicted labels with the input tags, we keep a binary vector for the active positions. 

\paragraph{Slot Filling as Token Classification} 

As in~\citep{devlin-etal-2019-bert}, we treat the slot filling as token classification, and we apply a shared layer on top of each token's representations to predict the tags.

Furthermore, we assemble the feature vector for the $i^{th}$ slot by concatenating together the predicted intent probabilities, the word features, and the contextual representation from the language model. Afterwards, we add a dropout followed by a linear projection to the proper number of slots:
\begin{equation}
    y^{i}_{s} = W_{s}[softmax(y_{int});f_{words}^{i};h_{LM}^{i}] + b_{s}
    \label{eq:slot:proj}
\end{equation}{}
 \vspace{-0.2in}
\subsection{Interaction and Learning}
To train the model, we use a joint loss function $\mathcal{L}_{joint}$ for the intent and for the slots. For both tasks, we apply cross-entropy over a softmax activation layer, except in the case of CRF tagging. In those experiments, the slot loss $\mathcal{L}_{slot}$ will be the negative log-likelihood (NLL) loss. Moreover, we introduce a new hyper-parameter $\gamma$ to balance the objectives of the two tasks (see Eq.~\ref{eq:joint:loss}). Finally, we propagate the loss from all the non-masked positions in the sequence, including word pieces, and special tokens ([CLS], $<$s$>$, etc.). Note that we do \textit{not} freeze any weights during fine-tuning. \rev{More details about the model can be found in  Appendix~\ref{sec:appx:model:details}.}
\vspace{-0.03in}
\begin{equation}
    \label{eq:joint:loss}
    \mathcal{L}_{joint} = \gamma * \mathcal{L}_{intent} + (1-\gamma) *  \mathcal{L}_{slot}
\end{equation}{}
\vspace{-0.2in}
\section{Experimental Setup}
\label{sec:setup}
\subsection{Dataset}
In our experiments, we use two publicly available datasets, the Airline Travel Information System (ATIS)~\citep{hemphill-etal-1990-atis}, and SNIPS~\citep{coucke2018snips}. The ATIS dataset contains transcripts from audio recordings of flight information requests, while the SNIPS dataset is gathered by a \rev{custom intent engine}
for personal voice assistants.  Albeit both are widely used in NLU benchmarks, ATIS is substantially smaller -- almost three times in terms of examples, and it contains s times less words. However, it has a richer set of labels, 21 intents and 120 slot categories, as opposed to the 7 intents and 72 slots in SNIPS. Another key difference is the diversity of domains -- ATIS has only utterances from the flight domain, while SNIPS covers various subjects, including entertainment, restaurant reservations, weather forecasts, etc. \rev{(see Table~\ref{tab:dataset:statistics})}
Furthermore, ATIS allows multiple intent labels. As they only form about 2\% of the data, we do not extend our model to multi-label classification. Yet, we add a new intent category for combinations seen in the training dataset, e.g.,~utterance with intents \textit{flight} and also \textit{airfare}, would be marked as \textit{airfare\#flight}.  
\begin{table}[t]
    \centering
    \small
    \begin{tabular}{l|c|c}
         \toprule
         & ATIS & SNIPS  \\
         \midrule
         Vocab Size & 722 & 11,241  \\
         Average Sentence Length & 11.28  & 9.05 \\
         \#Intents & 21 & 7  \\
         \#Slots & 120 & 72  \\
         \#Training Samples & 4,478 & 13,084 \\
         \#Dev Samples & 500 & 700 \\
         \#Test Samples & 893 & 700 \\
         \bottomrule
    \end{tabular}
    \caption{Statistics about the ATIS and SNIPS datasets.}
    \label{tab:dataset:statistics}
\end{table}{}

\subsection{Measures}
\label{sec:setup:metrics}
We evaluate our models with three well-established evaluation metrics.
The intent detection performance is measured in terms of accuracy. For the slot filling task, we use F1-score. Finally, the joint model is evaluated using sentence-level accuracy, i.e.,~proportion of examples in the corpus, whose intent and slots are both correctly predicted. Here, we must note that during evaluation we consider only the predictions for aligned words 
(we omit special tokens, %
and word pieces).

\newcommand{\twolineheader}[2]{\multicolumn{1}{c}{\centering #1 #2}}
\begin{table*}[ht!]
    \centering
    \setlength{\tabcolsep}{3pt} %
    \resizebox{\textwidth}{!}{%
    \begin{tabular}{l|ccc|ccc}
         \toprule
          & \multicolumn{3}{c}{ATIS} & \multicolumn{3}{c}{SNIPS} \\
         \midrule
         \bf Model  & \twolineheader{Intent}{(Acc)} & \twolineheader{Sent.}{(Acc)} & \twolineheader{Slot}{(F1)} & 
                      \twolineheader{Intent}{(Acc)} & \twolineheader{Sent.}{(Acc)} & \twolineheader{Slot}{(F1)} \\
         \hline
         \hline
         Joint Seq. \cite{hakkani-tr2016multi-domain} & \numOne{92.60}  & \numOne{80.70} & \numOne{94.30} & \numOne{96.90} & \numOne{73.20} & \numOne{87.30}   \\
         Atten.-Based \citep{liu2016attention} & \numOne{91.10} & \numOne{78.90} & \numOne{94.20} & \numOne{96.70} & \numOne{74.10} & \numOne{87.80}  \\
         Sloted-Gated \citep{goo-etal-2018-slot} & \numOne{95.41} & \numOne{83.73} & \numOne{95.42}  & \numOne{96.86} & \numOne{76.43}& \numOne{89.27} \\
         Capsule-NLU \citep{zhang-etal-2019-joint} & \numOne{95.00} & \numOne{83.40} & \numOne{95.20}  & \numOne{97.30} &  \numOne{80.90} & \numOne{91.80}   \\
         Interrelated SF-First \citep{e-etal-2019-novel} 
         & \numOne{97.76}  &  \numOne{86.79} & \numOne{95.75} & \numOne{97.43} & \numOne{80.57} & \numOne{91.43} \\
         
         Interrelated ID-First \citep{e-etal-2019-novel} 
         & \numOne{97.09} & \numOne{86.90} & \numOne{95.80} & \numOne{97.29} & \numOne{80.43} & \numOne{92.23} \\
         \hline
         \textit{BERT-Joint} & \numOne{97.42} & \numOne{87.57} & \numOne{95.74}  & \numOne{98.71} & \numOne{91.57} & \numOne{96.27} 
         \\
         \rev{\textit{RoBERTa-Joint}} & \numOne{97.42} & \numOne{87.23} & \numOne{95.32} & \numOne{98.71} & \numOne{90.71} & \numOne{95.85} \\
        \hline
        \hline
         \textit{Transformer-NLU:BERT} & \bf{\numOne{97.87}} & \bf{\numOne{88.69}} & \bf{\numOne{96.25}}  & \bf{\numOne{98.86}} & \numOne{91.86} & \bf{\numOne{96.57}}  
         \\
         \hline
         \rev{\textit{Transformer-NLU:RoBERTa}} & \numOne{97.76} & \numOne{87.91} & \numOne{95.65} & \bf{\numOne{98.86}} & \bf{\numOne{92.14}} & \numOne{96.35} \\
         \textit{Transformer-NLU:BERT w/o Slot Features} & \numOne{97.87} & \numOne{88.35} & \numOne{95.97}  & \numOne{98.86} & \numOne{91.57} & \numOne{96.25} \\
         \textit{Transformer-NLU:BERT w/ CRF} & \numOne{97.42}  & \numOne{88.26} & {\numOne{96.14}} & \numOne{98.57} & \numOne{92.00} & \numOne{96.54}  \\

         \bottomrule
    \end{tabular}
    }
    \caption{Intent detection and slot filling results on the SNIPS and the ATIS datasets. Highest results in each category are written in \textbf{bold}. \rev{Our models are shown in \emph{italic}; the non-italic models on top come from the literature.}
    }
    \label{tab:model:results}
\end{table*}{}

\subsection{Baselines} 
For our baseline models, we use BERT~\citep{devlin-etal-2019-bert} \rev{and RoBERTa~\citep{liu2019roberta}}, which we fine-tune. In particular, we train a linear layer over the pooled representation of the special [CLS] token to predict the intent. Moreover, we add a shared layer on top of the last hidden representations of the tokens in order to obtain a slot prediction \rev{(see Figure~\ref{subfig:base:bert})}.

\paragraph{BERT} 
For training the model, we follow the fine-tuning procedure proposed by~\citet{devlin-etal-2019-bert}. We train a linear layer over the pooled representation of the special [CLS] token to predict the utterance's intent. The latter is optimized during pre-training using the next sentence prediction (NSP) loss to encode the whole sentence. Moreover, we add a shared layer on top of the last hidden representations of the tokens in order to obtain a slot prediction. Both objectives are optimized using a cross-entropy loss.

\rev{
\paragraph{RoBERTa} 
This model follows the same training procedure as BERT, but drops the NSP task during pre-training. Still, the intent loss is attached to the special start token. %
}

\paragraph{State-of-the-art Models} 
We further compare our approach to some other benchmark models:
\begin{itemize}[leftmargin=*,nosep]
    \item Joint Seq. \citep{hakkani-tr2016multi-domain} uses a Recurrent Neural Network (RNN) to obtain hidden states for each token in the sequence for slot filling, and uses the last state to predict the intent.
    \item Atten.-Based \citep{liu2016attention} treats the slot filling task as a generative one, applying sequence-to-sequence RNN to label the input. Further, an attention weighted sum over the encoder's hidden states is used to detect the intent.
    \item Slotted-Gated \citep{goo-etal-2018-slot} introduces a special gated mechanism to an LSTM network, thus reinforcing the slot filling with the hidden representation used for the intent detection.
    \item Capsule-NLU \citep{zhang-etal-2019-joint} adopts Capsule Networks to exploit the semantic hierarchy between words, slots, and intents using dynamic routing-by-agreement schema.
    \item Interrelated \citep{e-etal-2019-novel} uses a Bidirectional LSTM  with attentive sub-networks for the slot and the intent modeling, and an interrelated mechanism to establish a direct connection between the two. SF (slot), and ID (intent) prefixes indicate which sub-network to execute first.
    \item BERT-Joint \citep{chen2019bert} used BERT with a token classification pipeline to jointly model the slot and the intent, with an additional CRF layer on top.\rev{\footnote{In terms of micro-average F1 for slot filling, \citet{chen2019bert} reported 96.1 on ATIS and 96.27 on SNIPS (per-token). For comparison, for our joint model, these scores are 98.1 and 97.9 (per-token); however, the correct scores for our model are actually 95.7 and 96.3 (per-slot).}
    However, they evaluated the slot filling task using per-token F1-score (micro averaging), rather than per-slot entry, as it is standard, which inflated their results. As their results are not comparable to the rest, we do not include them in our comparisons.}
\end{itemize}{}

\section{Model Details}
\label{sec:appx:model:details}

We use the PyTorch implementation of BERT from the Transformers library of~\citep{wolf-etal-2020-transformers} as a base for our models. We fine-tune all models for 50 epochs with hyper-parameters set as follows: batch size of 64 examples, maximum sequence length of 50 word pieces, dropout set to 0.1 (for both attentions and hidden layers), and weight decay of 0.01. For optimization, we use Adam with a learning rate of 8e-05, $\beta_1$ 0.9, $\beta_2$ 0.999,  $\epsilon$ 1e-06, and warm-up proportion of 0.1. 
Finally, in order to balance between the intent and the slot losses, we set the parameter $\gamma$ (Eq.~\ref{eq:joint:loss}) to 0.6, we test the range 0.4--0.8 with 0.1 increment. \rev{All the models use the same pre-processing, post-processing, and the standard for these tasks metrics}. 
In order to tackle the problem with random fluctuations for BERT\rev{/RoBERTa}, we ran the experiments three times and we used the best-performing model on the development set. We define the latter as the highest sum from all three measures described in Section~\ref{sec:setup:metrics}. 
All the above-mentioned hyper-parameter values were tuned on the development set, and then used for the final model on the test set. All models were trained on a single K80 GPU instance for around an hour.
\section{Experiments and Analysis}
\label{sec:results}

\noindent\paragraph{Evaluation Results}
Table~\ref{tab:model:results} presents quantitative evaluation results in terms of (\textit{i}) intent accuracy, (\textit{ii}) sentence accuracy, and (\textit{iii}) slot F1 (see Section~\ref{sec:setup:metrics}).
The first part of the table refers to previous work, whereas the second part presents our experiments and is separated with a double horizontal line. The evaluation results confirm that our model performs better then the current state-of-the-art (\rev{SOTA}). 

While, models become more accurate, the absolute difference between two experiments becomes smaller and smaller, thus a better measurement is needed. Hereby, we introduce a fine-grained measure, i.e.,~\textit{Relative Error Reduction} (RER) percentage, which is defined as the proportion of absolute error reduced by a $model_{a}$ compared to $model_{b}$.
\begin{equation}
    RER = 1 - \frac{Error_{model_a}}{Error_{model_b}}
    \label{eq:rer}
\end{equation}{} 
Table~\ref{tab:error:reduction} shows the error reduction by our model compared to the current \rev{SOTA}, and to a BERT-based baselines (\rev{see Appendix}~\ref{sec:appx:baselines}). Since there is no single best model from the SOTA, we take the per-column maximum among all, albeit they are not recorded in a single run. For the ATIS dataset, we see a reduction of 13.66\% (1.79 points absolute) for sentence accuracy, and 10.71\% (0.45 points absolute) for slot F1, but just 4.91\% for intent accuracy. 
Such a small improvement can be due to the quality of the dataset and to its size. 
For the SNIPS dataset, we see major increase in all measures and over 55\% error reduction. In absolute terms, we have 1.57 for intent, 10.96 for sentence, and 4.34 for slots. This effects cannot be only attributed to the better model (discussed in the analysis below), but also to the implicit information that BERT learned during its extensive pre-training. This is especially useful in the case of SNIPS, where fair amount of the slots in categories like \textit{SearchCreativeWork, SearchScreeningEvent, AddToPlaylist, PlayMusic} are names of movies, songs, artists, etc.

\begin{table}[t!]
    \centering
    \begin{tabular}{l|cc}
        \toprule
         Metric & \multicolumn{2}{c}{Relative Error Reduction} \\
         \midrule
          & \multicolumn{2}{c}{ATIS} \\ 
         Intent (Acc) & 4.91\% & 17.44\% \\
         Sent. (Acc) & 13.66\% & 11.43\% \\
         Slot (F1) & 10.71\% & 19.87\% \\
         & \multicolumn{2}{c}{SNIPS} \\ 
         Intent (Acc) & 55.64\% & 11.63 \% \\
         Sent. (Acc) & 57.38\% & 12.38\% \\
         Slot (F1) & 55.86\% & 17.35\% \\
         \midrule
         Transformer-NLU &  vs. SOTA &  vs. BERT \\
         \bottomrule
    \end{tabular}
    \caption{Relative error reduction (Eq.~\ref{eq:rer}) comparing \textit{Transformer-NLU:BERT} to the two baselines: \textit{{i})}~current \rev{SOTA} for each measure, and \textit{ii)}~conventionally fine-tuned BERT-Joint without the improvements.}
    \label{tab:error:reduction}
\end{table}{}

\rev{In addition to the aforementioned results, we also report the Transformer-NLU:BERT's (and BERT's) $\mu$ and $\sigma$
ATIS -- Intent $98.0\pm0.17$ (BERT $97.13\pm0.26$), Sentence $88.6\pm0.23$ (BERT $87.8\pm0$), Slot $96.3\pm0.06$ (BERT $96.0\pm0.14$); SNIPS -- Intent $98.6\pm0.14$ (BERT $98.42\pm0$), Sentence $92.0\pm0.17$ (BERT $91.8\pm0.19$), Slot $96.2\pm0.05$ (BERT $96.1\pm0.06$). 
The aforementioned results show that the mean scores of the models in the slot filling task are close, but the variance in Transformer-NLU is lower.
Further, we must note that these values are calculated over the best runs from each model re-training, and they are not achieved in a single run.

}

\paragraph{BERT Knowledge Analysis}
As we start to understand better BERT-based models~\citep{petroni2019language,rogers2020primer}, we observe some interesting phenomena.
BERT is trained on Wikipedia articles which allows it to learn implicit information about the world in addition to the language itself. 
Here, we evaluate how that knowledge reflects on the two NLU evaluation datasets. As a first step, we extract all the slot phrases from the training sets, i.e.,~all the words in the slot sequence. Next, we send the latter as a query to Wikipedia
and we collect the article titles. Then, we try to match the phrase with an extracted title. In order to reduce the false negatives, we normalize both texts (strip punctuation, replace digits with zeros, lower-case), allow difference of one character between the two, and finally if the title starts with the phrase, we count it as a match (e.g.,~\emph{Tampa} vs. \emph{Tampa, Florida}). Overall, 66\% of the slots in ATIS and 69\% in SNIPS matched a Wikipage title.

Next, we evaluate how much of that information is stored in the model by leveraging the standard masking mechanism used during pre-training. In particular, we split each slot in subwords, and then we replace them one by one sequentially with the special [MASK] token. We then sort the predictions for that position by probability and we take the rank of the true word. Finally, we calculate the mean reciprocal rank (MRR) over all the aforementioned ranks: 0.64 for ATIS,
and 0.36 for SNIPS.
We must note that the BERT's dictionary contains 32K pieces, and the expected uniform MRR is $\sim$1/16,000. Below, we present two examples to illustrate both high- and low-ranked predictions.\\
\textbf{High ranked:} \emph{play the album jack takes the floor by tom le [MASK] on netflix},
here the 
top predictions are: [{\textbf{\#\#hrer}, \textit{\#\#rner, \#\#mmon, \#\#hr}}],
and the correct token is \rev{ranked} with the highest probability. \\
\textbf{Low ranked:} \emph{play some hong jun [MASK]},
here the model's top guesses are mostly punctuation, and general words such as [\textit{to, ;, \#\#s, and}]. The correct token \textit{\#\#yang} is at position 3,036, which indicates that this term is challenging.

In SNIPS, we can see that types such as \textit{track, movie\_name, entry\_name, album} have very high MRR (0.4), and ones that require numerical value, or are not part of well-known named entities such as object\_part\_of\_series\_type are the lowest (under 0.1). The same in ATIS for country\_name (8e\mbox{-}3), restriction\_code (4e\mbox{-}3), meal (4e\mbox{-}3), in contrast to airline\_code (0.45), transport\_type (0.42), etc. However, ATIS in general does not require such task-specific knowledge, and its MRR is way higher in general, which is reflected by the overall improvement compared to the baseline models.

\noindent\paragraph{Transformer-NLU Analysis}
We dissect the proposed model by adding or removing prominent components to outline their contributions. The results are shown in the second part of Table~\ref{tab:model:results}. 
First, we compare the results of \textit{BERT-Joint} and the enriched model \textit{Transformer-NLU:BERT}. We can see a notable reduction of the intent classification error by 17.44\% and 11.63\% for the ATIS and the SNIPS dataset, respectively. Furthermore, we see a 19.87\% (ATIS) and 17.35\% (SNIPS) error reduction in slot's F1, and 11.43\% (ATIS) and 11.63\% (SNIPS) for sentence accuracy.
\rev{We also try RoBERTa as a backbone to our model: while we still see the positive effect of the proposed architecture, the overall results are slightly worse. We attribute this to the different set of pre-training data (CommonCrawl vs. Wikipedia). We further focus our analysis on BERT-based models, since they performed better than RoBERTa-based ones.}

Next, we remove the additional slot features~-- predicted intent, word casing, and named entities. The results are shown as {Transformer-NLU:BERT w/o Slot Features}. As expected, the intent accuracy remains unchanged for both datasets, since we retain the pooling attention layer, while the F1-score for the slots decreases. For SNIPS, the model achieved the same score as for \textit{BERT-Joint}, while for ATIS it was within 0.2 points absolute.

Finally, we added a CRF layer on top of the slot network, since it had shown positive effects in earlier studies~\citep{xu2013convolutional,huang2015bidirectional,liu2016attention,e-etal-2019-novel}. We denote the experiment as \textit{Transformer-NLU:BERT w/ CRF}. However, in our case it did not yield the expected improvement. The results for slot filling are close to the highest recorded, while a drastic drop in intent detection accuracy is observed, i.e.,~-17.44\% for ATIS, and -20.28\% for SNIPS. We attribute this degradation to the large gradients from the NLL loss. The effect is even stronger in the case of smaller datasets, making the optimization unstable for parameter-rich models such as BERT.
We tried to mitigate this issue by increasing the $\gamma$ hyper-parameter, effectively reducing the contribution of the slot's loss $\mathcal{L}_{slot}$ to the total, which in turn harmed the slot's F1. \rev{Moreover, the model does swap interchangeable slots, rather than the \textit{B-} and \textit{I-} prefixes, or slots unrelated to the intent (see the \textbf{Error Analysis} below).}
\noindent\paragraph{Intent Pooling Attention Visualization}
\begin{figure}[t!]
    \centering
    \includegraphics[width=0.38\textwidth, height=3.2cm]{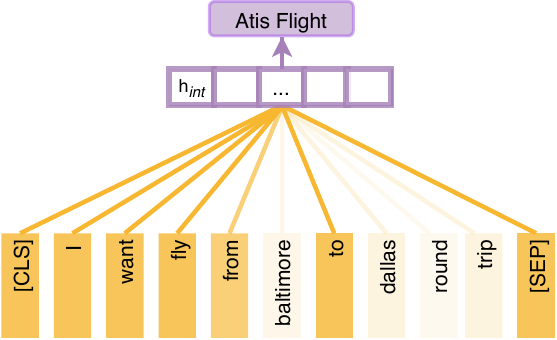}

    \caption{Intent pooling attention weight for atis\_flight. The thicker the line, the higher the attention.}
    \label{fig:intent:attn:atis}
\end{figure}{}

Next, we visualize the learned attention weights on Figure~\ref{fig:intent:attn:atis}. It presents a request from the ATIS dataset: \textit{i want fly from baltimore to dallas round trip}. The utterance's intent is marked as {\textit{atis\_flight}}, and we can see that the attention successfully picked the key tokens, i.e.,~\textit{I}, \textit{want}, \textit{fly}, \textit{from}, and \textit{to}, whereas supplementary words such as names, locations, dates, etc. have less contribution. Moreover, when trained on the ATIS dataset, the layer tends to set the weights in the two extremes --- equally high for important tokens, and towards zero for the rest. We attribute this to the limited domain and vocabulary.

Finally, we let the pooling attention layer consider the special tokens marking the start and the end ([CLS], and [SEP]) of a sequence, since they are expected to learn semantic sentence-level representations from the penultimate layer. The model assigns high attention weights to both.

\rev{
\paragraph{Error Analysis} 
First, we compare the performance of our method (\textit{Transformer-NLU}) to \textit{BERT-Joint (BERT)}. In the intent detection task, the largest improvement (over BERT) comes from examples with slots, indicative for a given intent. This suggests that the model successfully uses the slot information gathered by the pooling attention layer. For the following groups, this is most prominent:
(\textit{i})~examples with multi-label intents, e.g.,~\textit{atis\_airline\#atis\_flight\_no} -- \textit{``i need \textbf{flight numbers} and \textbf{airlines} \dots''}, where \textit{BERT} predicted \textit{atis\_flight\_no}; 
(\textit{ii})~examples containing distinctive words for another intent class -- \textit{``Give me \textbf{meal} flights ...''}, \textit{atis\_flight $\to$ meal~(BERT)}, \textit{``I need a \textbf{table} \dots when it is chiller''}, \textit{GetWeather $\to$ BookRestaurant~(BERT)}. For all the aforementioned examples, both models filled the slots correctly, but only \textit{Transformer-NLU} captured the correct intent. Moreover, we see a positive effect in detecting \textit{SearchCreativeWork} and \textit{SearchScreeningEvent}, while BERT tends to wrongly fill the slots, and thus swaps the two intents, e.g.,~\textit{``find \textbf{heat wave}''}, or \textit{``find \textbf{now and forever}''}. Finally, we see an additional improvement for \textit{AddToPlaylist} and \textit{atis\_ground\_fare}.

Next, we compare
the two models on the slot filling task. As expected, the newly proposed model performs better, when the curated features capture some interesting phenomena. We observe that, when filling code slots (\textbf{airport, meal, airfare}) -- \textit{``what does \dots code \textbf{bh} mean''}, artists, albums, movies, object names -- \textit{\textbf{dwele, nyoil, turk}} (\textit{artist $\to$ entity\_name (BERT)}), locations -- \textit{``\dots between milwaukee and \textbf{indiana}''}, \textit{state $\to$ city (BERT)}; BERT confuses \textit{\textbf{mango} (city)} with the fruit (cuisine); \textit{``new york city \textbf{area}''} \textit{O $\to$ city (BERT)} and time-related ones -- \textit{\textbf{afternoon, late night, a.m.}}. 

Finally, we discuss the errors of \textit{Transformer-NLU} in general. For the ATIS dataset, 50\% of the wrong intents come from multi-label cases (35\% with two labels, and 15\% with three), 31\% \textit{atis\_flight} -- \textit{``how many \textbf{flights} does \dots \textbf{/have to/leave} \dots''} ($\to$ atis\_quantity), 11\% \textit{atis\_city} -- \textit{list la} ($\to$ atis\_abbreviation), and the others are mistakes in \textit{atis\_aircraft}. For the slots, 50\% of the errors come from tags that can have a \textit{fromloc/toloc} prefix, e.g.,~\textit{city, airport\_code, airport\_name, etc.}, another 20\% are time-related (\textit{arrive\_date, return\_date}), and filled slots without tag 7\%.
The errors by the model for the SNIPS datasets are as follows: mislabeled intents \textit{PlayMusic} 11\%, \textit{SearchCreativeWork} 22\%, \textit{SearchScreeningEvent} 67\%, slots -- \textit{movie\_name} 19\%, \textit{object\_name} 16\%,  \textit{playlist} 9\%,  track 9\%, entity\_name 5\%, \textit{album} 4\%, \textit{timeRange} 4\%, \textit{served\_dish} 2\%, filled slots without tag 19\%. The model misses 9\% (ATIS) and 17\% (SNIPS) of all the slots that should be filled. This is expected since SNIPS' slots have a larger dictionary (11K words), with a large proportion of the slots being names, and often including prepositions, e.g.,~\textit{``\dots trailer of \textbf{the multiversity}''}.

}

\section{Related Work}
\label{sec:relatedwork}

\paragraph{Intent Classification} Several approaches focus only on the utterance's intent, and omit slot information, i.e.,~\citet{Hu:2009:UUQ:1526709.1526773} maps each intent domain and user's queries into Wikipedia representation space,~\citet{kim2017two} and~\citet{xu2013exploiting} use log-linear models with multiple-stages and word features, while~\citep{ravuri2015recurrent} investigate word and character n-gram language models based on Recurrent Neural Networks and LSTMs~\citep{Hochreiter:1997:LSM:1246443.1246450}, \citet{xia-etal-2018-zero} proposed a zero-shot transfer thought Capsule Networks~\citep{sabour2017dynamic} and semantic features for detecting user intents where no labeled data exists. Moreover, some work extend the problem to a multi-class multi-label one~\citep{xu2013exploiting,kim2017two,gangadharaiah-narayanaswamy-2019-joint}. 

\noindent\paragraph{Slot Filling}
Before the rise of deep learning, sequential models such as Maximum Entropy Markov model (MEMM) \citep{toutanvoa-manning-2000-enriching,mccallum2000maximum}, and Conditional Random Fields (CRF)~\cite{Lafferty:2001:CRF:645530.655813,jeong2008triangular} were the state-of-the-art choice.
Recently, several combinations thereof and different neural network architecture were proposed~\citep{xu2013convolutional,huang2015bidirectional,e-etal-2019-novel}. However, a steer away from sequential models is observed in favor of self-attentive ones such as the
Transformer~\citep{devlin-etal-2019-bert,liu2019roberta,radford2018improving,radford2019language}. They compose a contextualized representation of both the sentences, and each word, though a sequence of intermediate non-linear hidden layers, %
usually followed by a projection layer in order to obtain per-token tags. Such models had been successfully applied to closely related tasks, e.g.,~NER~\citep{devlin-etal-2019-bert}, POS tagging~\citep{tsai-etal-2019-small}, etc.

Approaches modeling the intent or the slot as independent of each other suffer from  uncertainty propagation due the absence of shared knowledge between the tasks. To overcome this limitation, we learn both tasks using a joint model.

\paragraph{Joint Models}

Given the correlation between the intent and word-level slot tags, it is natural to train them jointly. \citet{xu2013convolutional} introduced a shared intent and slot hidden state Convolutional Neural Network (CNN)  followed by a globally normalized CRF (TriCRF) for sequence tagging. Since then, Recurrent Neural Networks have been dominating, e.g.,~\citet{hakkani-tr2016multi-domain} used bidirectional LSTMs for slot filling and the last hidden state for intent classification, \citet{liu2016attention} introduced shared attention weights between the slot and the intent layer. \citet{goo-etal-2018-slot} integrated the intent via a gating mechanism into the context vector of LSTM cells used for slot filling. 

\citet{e-etal-2019-novel} introduced a bidirectional interrelated model, using an iterative mechanism to correct the predicted intent and slot by multiple step refinement. %
\citet{zhang-etal-2019-joint} tried to exploit the semantic hierarchical relationship between words, slots, and intent via a dynamic
\rev{routing by agreement} 
schema in a \rev{Capsule Net}~\citep{sabour2017dynamic}.

Here, we used a pre-trained Transformer, and instead of depending only on the language model's hidden state to learn the interaction between the slot and the intent, we fuse the two tasks together. 
Moreover, we leverage information from external sources: \textit{(i)}~from explicit NER and true case annotations, \textit{(ii)}~from implicit information learned by the language model during its extensive pre-training.
\section{Conclusion and Future Work}
\label{sec:conclusions}

We studied the two main challenges in natural language understanding, i.e.,~intent detection and slot filling. In particular, we proposed an enriched pre-trained language model to jointly model the two tasks, i.e.,~\textit{Transformer-NLU}. We designed a pooling attention layer in order to obtain intent representation beyond just the pooled one from the special start token. Further, we reinforced the slot filling with word-specific features, and the predicted intent distribution. 
Our experiments on two standard datasets showed that Transformer-NLU outperforms other alternatives for all standard measures used to evaluate NLU tasks. \rev{We found that the use of RoBERTa and} adding a CRF layer on top of the slot filling network did not help. Finally, the Transformer-NLU:BERT achieved intent accuracy of 97.87 (ATIS) and 98.86 (SNIPS). Or as a relative error reduction -- almost 5\% for ATIS, and over 55\% for SNIPS, compared to the state-of-the-art. In terms of \rev{slot's}
F1, we scored 96.25 (+13.66\%) for ATIS, and 96.57 (+55.86\%) for SNIPS.

In future work, we plan to investigate the natural hierarchy of the slots, e.g.,~\textit{B-toloc.city} can be split into \textit{B}, \textit{toloc}, and \textit{city}. We further want to try a better named entity recognition framework such as FLAIR~\citep{akbik-etal-2019-flair,akbik-etal-2019-pooled}.
\rev{
\section*{Ethics and Broader Impact}

Our intent pooling mechanism, as well as the features we introduced, are potentially applicable to other semantic parsing and sequence labeling tasks. They increase the model's size by just few tens of thousands of parameters, which is very efficient in comparison to modern NLP models, which have millions or even billions of parameters. 

On the down side, we would like to warn about the potential biases in the data used for training Transformers such as BERT and RoBERTa, as well as in the ATIS and the SNIPS datasets. Moreover, the use of large-scale Transformers and GPUs could contribute to global warming.
}

\bibliographystyle{acl_natbib}
\bibliography{custom}

\begin{thebibliography}{43}
\expandafter\ifx\csname natexlab\endcsname\relax\def\natexlab#1{#1}\fi

\bibitem[{Akbik et~al.(2019{\natexlab{a}})Akbik, Bergmann, Blythe, Rasul,
  Schweter, and Vollgraf}]{akbik-etal-2019-flair}
Alan Akbik, Tanja Bergmann, Duncan Blythe, Kashif Rasul, Stefan Schweter, and
  Roland Vollgraf. 2019{\natexlab{a}}.
\newblock \href {https://doi.org/10.18653/v1/N19-4010} {{FLAIR}: An easy-to-use
  framework for state-of-the-art {NLP}}.
\newblock In \emph{Proceedings of the 2019 Conference of the North {A}merican
  Chapter of the Association for Computational Linguistics (Demonstrations)},
  pages 54--59, Minneapolis, Minnesota. Association for Computational
  Linguistics.

\bibitem[{Akbik et~al.(2019{\natexlab{b}})Akbik, Bergmann, and
  Vollgraf}]{akbik-etal-2019-pooled}
Alan Akbik, Tanja Bergmann, and Roland Vollgraf. 2019{\natexlab{b}}.
\newblock \href {https://doi.org/10.18653/v1/N19-1078} {Pooled contextualized
  embeddings for named entity recognition}.
\newblock In \emph{Proceedings of the 2019 Conference of the North {A}merican
  Chapter of the Association for Computational Linguistics: Human Language
  Technologies, Volume 1 (Long and Short Papers)}, pages 724--728, Minneapolis,
  Minnesota. Association for Computational Linguistics.

\bibitem[{Chen et~al.(2019)Chen, Zhuo, and Wang}]{chen2019bert}
Qian Chen, Zhu Zhuo, and Wen Wang. 2019.
\newblock {BERT} for joint intent classification and slot filling.
\newblock \emph{arXiv preprint arXiv:1902.10909}.

\bibitem[{Conneau and Lample(2019)}]{lample2019cross}
Alexis Conneau and Guillaume Lample. 2019.
\newblock \href
  {https://proceedings.neurips.cc/paper/2019/hash/c04c19c2c2474dbf5f7ac4372c5b9af1-Abstract.html}
  {Cross-lingual language model pretraining}.
\newblock In \emph{Advances in Neural Information Processing Systems 32: Annual
  Conference on Neural Information Processing Systems 2019, NeurIPS 2019,
  December 8-14, 2019, Vancouver, BC, Canada}, pages 7057--7067.

\bibitem[{Coucke et~al.(2018)Coucke, Saade, Ball, Bluche, Caulier, Leroy,
  Doumouro, Gisselbrecht, Caltagirone, Lavril et~al.}]{coucke2018snips}
Alice Coucke, Alaa Saade, Adrien Ball, Th{\'e}odore Bluche, Alexandre Caulier,
  David Leroy, Cl{\'e}ment Doumouro, Thibault Gisselbrecht, Francesco
  Caltagirone, Thibaut Lavril, et~al. 2018.
\newblock {S}nips voice platform: an embedded spoken language understanding
  system for private-by-design voice interfaces.
\newblock \emph{arXiv:1805.10190}.

\bibitem[{Devlin et~al.(2019)Devlin, Chang, Lee, and
  Toutanova}]{devlin-etal-2019-bert}
Jacob Devlin, Ming-Wei Chang, Kenton Lee, and Kristina Toutanova. 2019.
\newblock \href {https://doi.org/10.18653/v1/N19-1423} {{BERT}: Pre-training of
  deep bidirectional transformers for language understanding}.
\newblock In \emph{Proceedings of the 2019 Conference of the North {A}merican
  Chapter of the Association for Computational Linguistics: Human Language
  Technologies, Volume 1 (Long and Short Papers)}, pages 4171--4186,
  Minneapolis, Minnesota. Association for Computational Linguistics.

\bibitem[{E et~al.(2019)E, Niu, Chen, and Song}]{e-etal-2019-novel}
Haihong E, Peiqing Niu, Zhongfu Chen, and Meina Song. 2019.
\newblock \href {https://doi.org/10.18653/v1/P19-1544} {A novel bi-directional
  interrelated model for joint intent detection and slot filling}.
\newblock In \emph{Proceedings of the 57th Annual Meeting of the Association
  for Computational Linguistics}, pages 5467--5471, Florence, Italy.
  Association for Computational Linguistics.

\bibitem[{Gangadharaiah and
  Narayanaswamy(2019)}]{gangadharaiah-narayanaswamy-2019-joint}
Rashmi Gangadharaiah and Balakrishnan Narayanaswamy. 2019.
\newblock \href {https://doi.org/10.18653/v1/N19-1055} {Joint multiple intent
  detection and slot labeling for goal-oriented dialog}.
\newblock In \emph{Proceedings of the 2019 Conference of the North {A}merican
  Chapter of the Association for Computational Linguistics: Human Language
  Technologies, Volume 1 (Long and Short Papers)}, pages 564--569, Minneapolis,
  Minnesota. Association for Computational Linguistics.

\bibitem[{Goo et~al.(2018)Goo, Gao, Hsu, Huo, Chen, Hsu, and
  Chen}]{goo-etal-2018-slot}
Chih-Wen Goo, Guang Gao, Yun-Kai Hsu, Chih-Li Huo, Tsung-Chieh Chen, Keng-Wei
  Hsu, and Yun-Nung Chen. 2018.
\newblock \href {https://doi.org/10.18653/v1/N18-2118} {Slot-gated modeling for
  joint slot filling and intent prediction}.
\newblock In \emph{Proceedings of the 2018 Conference of the North {A}merican
  Chapter of the Association for Computational Linguistics: Human Language
  Technologies, Volume 2 (Short Papers)}, pages 753--757, New Orleans,
  Louisiana. Association for Computational Linguistics.

\bibitem[{Hakkani-T{\"u}r et~al.(2016)Hakkani-T{\"u}r, Tur, Celikyilmaz, Chen,
  Gao, Deng, and Wang}]{hakkani-tr2016multi-domain}
Dilek Hakkani-T{\"u}r, Gokhan Tur, Asli Celikyilmaz, Yun-Nung~Vivian Chen,
  Jianfeng Gao, Li~Deng, and Ye-Yi Wang. 2016.
\newblock Multi-domain joint semantic frame parsing using bi-directional
  {RNN-LSTM}.
\newblock In \emph{Proceedings of The 17th Annual Meeting of the International
  Speech Communication Association}, INTERSPEECH~'16.

\bibitem[{Hemphill et~al.(1990)Hemphill, Godfrey, and
  Doddington}]{hemphill-etal-1990-atis}
Charles~T. Hemphill, John~J. Godfrey, and George~R. Doddington. 1990.
\newblock \href {https://www.aclweb.org/anthology/H90-1021} {The {ATIS} spoken
  language systems pilot corpus}.
\newblock In \emph{Speech and Natural Language: Proceedings of a Workshop Held
  at Hidden Valley, {P}ennsylvania, June 24-27,1990}.

\bibitem[{Hochreiter and
  Schmidhuber(1997)}]{Hochreiter:1997:LSM:1246443.1246450}
Sepp Hochreiter and J\"{u}rgen Schmidhuber. 1997.
\newblock Long short-term memory.
\newblock \emph{Neural Computation}, 9(8):1735--1780.

\bibitem[{Hu et~al.(2009)Hu, Wang, Lochovsky, Sun, and
  Chen}]{Hu:2009:UUQ:1526709.1526773}
Jian Hu, Gang Wang, Frederick~H. Lochovsky, Jian{-}Tao Sun, and Zheng Chen.
  2009.
\newblock \href {https://doi.org/10.1145/1526709.1526773} {Understanding user's
  query intent with wikipedia}.
\newblock In \emph{Proceedings of the 18th International Conference on World
  Wide Web, {WWW} 2009, Madrid, Spain, April 20-24, 2009}, pages 471--480.
  {ACM}.

\bibitem[{Huang et~al.(2015)Huang, Xu, and Yu}]{huang2015bidirectional}
Zhiheng Huang, Wei Xu, and Kai Yu. 2015.
\newblock Bidirectional {LSTM-CRF} models for sequence tagging.
\newblock \emph{arXiv:1508.01991}.

\bibitem[{Jeong and Lee(2008)}]{jeong2008triangular}
Minwoo Jeong and Gary~Geunbae Lee. 2008.
\newblock Triangular-chain conditional random fields.
\newblock \emph{IEEE Transactions on Audio, Speech, and Language Processing},
  16(7):1287--1302.

\bibitem[{Joshi et~al.(2020)Joshi, Chen, Liu, Weld, Zettlemoyer, and
  Levy}]{joshi2019spanbert}
Mandar Joshi, Danqi Chen, Yinhan Liu, Daniel~S. Weld, Luke Zettlemoyer, and
  Omer Levy. 2020.
\newblock \href {https://doi.org/10.1162/tacl_a_00300} {{S}pan{BERT}: Improving
  pre-training by representing and predicting spans}.
\newblock \emph{Transactions of the Association for Computational Linguistics},
  8:64--77.

\bibitem[{Kim et~al.(2017)Kim, Ryu, and Lee}]{kim2017two}
Byeongchang Kim, Seonghan Ryu, and Gary~Geunbae Lee. 2017.
\newblock Two-stage multi-intent detection for spoken language understanding.
\newblock \emph{Multimedia Tools and Applications}, 76(9):11377--11390.

\bibitem[{Kudo and Richardson(2018)}]{kudo-richardson-2018-sentencepiece}
Taku Kudo and John Richardson. 2018.
\newblock \href {https://doi.org/10.18653/v1/D18-2012} {{S}entence{P}iece: A
  simple and language independent subword tokenizer and detokenizer for neural
  text processing}.
\newblock In \emph{Proceedings of the 2018 Conference on Empirical Methods in
  Natural Language Processing: System Demonstrations}, pages 66--71, Brussels,
  Belgium. Association for Computational Linguistics.

\bibitem[{Lafferty et~al.(2001)Lafferty, McCallum, and
  Pereira}]{Lafferty:2001:CRF:645530.655813}
John~D. Lafferty, Andrew McCallum, and Fernando C.~N. Pereira. 2001.
\newblock Conditional random fields: Probabilistic models for segmenting and
  labeling sequence data.
\newblock In \emph{Proceedings of the Eighteenth International Conference on
  Machine Learning {(ICML} 2001), Williams College, Williamstown, MA, USA, June
  28 - July 1, 2001}, pages 282--289. Morgan Kaufmann.

\bibitem[{Lan et~al.(2020)Lan, Chen, Goodman, Gimpel, Sharma, and
  Soricut}]{lan2019albert}
Zhenzhong Lan, Mingda Chen, Sebastian Goodman, Kevin Gimpel, Piyush Sharma, and
  Radu Soricut. 2020.
\newblock \href {https://openreview.net/forum?id=H1eA7AEtvS} {{ALBERT:} {A}
  lite {BERT} for self-supervised learning of language representations}.
\newblock In \emph{8th International Conference on Learning Representations,
  {ICLR} 2020, Addis Ababa, Ethiopia, April 26-30, 2020}. OpenReview.net.

\bibitem[{Liu and Lane(2016)}]{liu2016attention}
Bing Liu and Ian Lane. 2016.
\newblock Attention-based recurrent neural network models for joint intent
  detection and slot filling.
\newblock In \emph{Proceedings of The 17th Annual Meeting of the International
  Speech Communication Association}, INTERSPEECH~'16, pages 685--689.

\bibitem[{Liu et~al.(2019)Liu, Ott, Goyal, Du, Joshi, Chen, Levy, Lewis,
  Zettlemoyer, and Stoyanov}]{liu2019roberta}
Yinhan Liu, Myle Ott, Naman Goyal, Jingfei Du, Mandar Joshi, Danqi Chen, Omer
  Levy, Mike Lewis, Luke Zettlemoyer, and Veselin Stoyanov. 2019.
\newblock {RoBERTa}: A robustly optimized {BERT} pretraining approach.
\newblock \emph{arXiv:1907.11692}.

\bibitem[{Luong et~al.(2015)Luong, Pham, and
  Manning}]{luong-etal-2015-effective}
Thang Luong, Hieu Pham, and Christopher~D. Manning. 2015.
\newblock \href {https://doi.org/10.18653/v1/D15-1166} {Effective approaches to
  attention-based neural machine translation}.
\newblock In \emph{Proceedings of the 2015 Conference on Empirical Methods in
  Natural Language Processing}, pages 1412--1421, Lisbon, Portugal. Association
  for Computational Linguistics.

\bibitem[{Manning et~al.(2014)Manning, Surdeanu, Bauer, Finkel, Bethard, and
  McClosky}]{manning-EtAl:2014:P14-5}
Christopher Manning, Mihai Surdeanu, John Bauer, Jenny Finkel, Steven Bethard,
  and David McClosky. 2014.
\newblock \href {https://doi.org/10.3115/v1/P14-5010} {The {S}tanford
  {C}ore{NLP} natural language processing toolkit}.
\newblock In \emph{Proceedings of 52nd Annual Meeting of the Association for
  Computational Linguistics: System Demonstrations}, pages 55--60, Baltimore,
  Maryland. Association for Computational Linguistics.

\bibitem[{McCallum et~al.(2000)McCallum, Freitag, and
  Pereira}]{mccallum2000maximum}
Andrew McCallum, Dayne Freitag, and Fernando C.~N. Pereira. 2000.
\newblock Maximum entropy markov models for information extraction and
  segmentation.
\newblock In \emph{Proceedings of the Seventeenth International Conference on
  Machine Learning {(ICML} 2000), Stanford University, Stanford, CA, USA, June
  29 - July 2, 2000}, pages 591--598. Morgan Kaufmann.

\bibitem[{Petroni et~al.(2019)Petroni, Rockt{\"a}schel, Riedel, Lewis, Bakhtin,
  Wu, and Miller}]{petroni2019language}
Fabio Petroni, Tim Rockt{\"a}schel, Sebastian Riedel, Patrick Lewis, Anton
  Bakhtin, Yuxiang Wu, and Alexander Miller. 2019.
\newblock \href {https://doi.org/10.18653/v1/D19-1250} {Language models as
  knowledge bases?}
\newblock In \emph{Proceedings of the 2019 Conference on Empirical Methods in
  Natural Language Processing and the 9th International Joint Conference on
  Natural Language Processing (EMNLP-IJCNLP)}, pages 2463--2473, Hong Kong,
  China. Association for Computational Linguistics.

\bibitem[{Radford et~al.(2018)Radford, Narasimhan, Salimans, and
  Sutskever}]{radford2018improving}
Alec Radford, Karthik Narasimhan, Tim Salimans, and Ilya Sutskever. 2018.
\newblock Improving language understanding by generative pre-training.

\bibitem[{Radford et~al.(2019)Radford, Wu, Child, Luan, Amodei, and
  Sutskever}]{radford2019language}
Alec Radford, Jeffrey Wu, Rewon Child, David Luan, Dario Amodei, and Ilya
  Sutskever. 2019.
\newblock Language models are unsupervised multitask learners.
\newblock \emph{OpenAI Blog}.

\bibitem[{Ravuri and Stolcke(2015)}]{ravuri2015recurrent}
Suman Ravuri and Andreas Stolcke. 2015.
\newblock Recurrent neural network and {LSTM} models for lexical utterance
  classification.
\newblock In \emph{Sixteenth Annual Conference of the International Speech
  Communication Association}.

\bibitem[{Rogers et~al.(2020)Rogers, Kovaleva, and
  Rumshisky}]{rogers2020primer}
Anna Rogers, Olga Kovaleva, and Anna Rumshisky. 2020.
\newblock \href {https://doi.org/10.1162/tacl_a_00349} {A primer in
  {BERT}ology: What we know about how {BERT} works}.
\newblock \emph{Transactions of the Association for Computational Linguistics},
  8:842--866.

\bibitem[{Sabour et~al.(2017)Sabour, Frosst, and Hinton}]{sabour2017dynamic}
Sara Sabour, Nicholas Frosst, and Geoffrey~E. Hinton. 2017.
\newblock \href
  {https://proceedings.neurips.cc/paper/2017/hash/2cad8fa47bbef282badbb8de5374b894-Abstract.html}
  {Dynamic routing between capsules}.
\newblock In \emph{Advances in Neural Information Processing Systems 30: Annual
  Conference on Neural Information Processing Systems 2017, December 4-9, 2017,
  Long Beach, CA, {USA}}, pages 3856--3866.

\bibitem[{Sennrich et~al.(2016)Sennrich, Haddow, and
  Birch}]{sennrich-etal-2016-neural}
Rico Sennrich, Barry Haddow, and Alexandra Birch. 2016.
\newblock \href {https://doi.org/10.18653/v1/P16-1162} {Neural machine
  translation of rare words with subword units}.
\newblock In \emph{Proceedings of the 54th Annual Meeting of the Association
  for Computational Linguistics (Volume 1: Long Papers)}, pages 1715--1725,
  Berlin, Germany. Association for Computational Linguistics.

\bibitem[{Toutanvoa and Manning(2000)}]{toutanvoa-manning-2000-enriching}
Kristina Toutanvoa and Christopher~D. Manning. 2000.
\newblock \href {https://doi.org/10.3115/1117794.1117802} {Enriching the
  knowledge sources used in a maximum entropy part-of-speech tagger}.
\newblock In \emph{2000 Joint {SIGDAT} Conference on Empirical Methods in
  Natural Language Processing and Very Large Corpora}, pages 63--70, Hong Kong,
  China. Association for Computational Linguistics.

\bibitem[{Tsai et~al.(2019)Tsai, Riesa, Johnson, Arivazhagan, Li, and
  Archer}]{tsai-etal-2019-small}
Henry Tsai, Jason Riesa, Melvin Johnson, Naveen Arivazhagan, Xin Li, and Amelia
  Archer. 2019.
\newblock \href {https://doi.org/10.18653/v1/D19-1374} {Small and practical
  {BERT} models for sequence labeling}.
\newblock In \emph{Proceedings of the 2019 Conference on Empirical Methods in
  Natural Language Processing and the 9th International Joint Conference on
  Natural Language Processing (EMNLP-IJCNLP)}, pages 3632--3636, Hong Kong,
  China. Association for Computational Linguistics.

\bibitem[{Tur et~al.(2010)Tur, Hakkani-T{\"u}r, and Heck}]{tur2010left}
Gokhan Tur, Dilek Hakkani-T{\"u}r, and Larry Heck. 2010.
\newblock \href {https://doi.org/10.1109/SLT.2010.5700816} {What is left to be
  understood in {ATIS}?}
\newblock In \emph{2010 IEEE Spoken Language Technology Workshop}, pages
  19--24. IEEE.

\bibitem[{Vaswani et~al.(2017)Vaswani, Shazeer, Parmar, Uszkoreit, Jones,
  Gomez, Kaiser, and Polosukhin}]{NIPS2017_7181:transformer}
Ashish Vaswani, Noam Shazeer, Niki Parmar, Jakob Uszkoreit, Llion Jones,
  Aidan~N. Gomez, Lukasz Kaiser, and Illia Polosukhin. 2017.
\newblock \href
  {https://proceedings.neurips.cc/paper/2017/hash/3f5ee243547dee91fbd053c1c4a845aa-Abstract.html}
  {Attention is all you need}.
\newblock In \emph{Advances in Neural Information Processing Systems 30: Annual
  Conference on Neural Information Processing Systems 2017, December 4-9, 2017,
  Long Beach, CA, {USA}}, pages 5998--6008.

\bibitem[{Wolf et~al.(2020)Wolf, Debut, Sanh, Chaumond, Delangue, Moi, Cistac,
  Rault, Louf, Funtowicz, Davison, Shleifer, von Platen, Ma, Jernite, Plu, Xu,
  Le~Scao, Gugger, Drame, Lhoest, and Rush}]{wolf-etal-2020-transformers}
Thomas Wolf, Lysandre Debut, Victor Sanh, Julien Chaumond, Clement Delangue,
  Anthony Moi, Pierric Cistac, Tim Rault, Remi Louf, Morgan Funtowicz, Joe
  Davison, Sam Shleifer, Patrick von Platen, Clara Ma, Yacine Jernite, Julien
  Plu, Canwen Xu, Teven Le~Scao, Sylvain Gugger, Mariama Drame, Quentin Lhoest,
  and Alexander Rush. 2020.
\newblock \href {https://doi.org/10.18653/v1/2020.emnlp-demos.6} {Transformers:
  State-of-the-art natural language processing}.
\newblock In \emph{Proceedings of the 2020 Conference on Empirical Methods in
  Natural Language Processing: System Demonstrations}, pages 38--45, Online.
  Association for Computational Linguistics.

\bibitem[{Wu et~al.(2016)Wu, Schuster, Chen, Le, Norouzi, Macherey, Krikun,
  Cao, Gao, Macherey et~al.}]{wu2016google}
Yonghui Wu, Mike Schuster, Zhifeng Chen, Quoc~V Le, Mohammad Norouzi, Wolfgang
  Macherey, Maxim Krikun, Yuan Cao, Qin Gao, Klaus Macherey, et~al. 2016.
\newblock {G}oogle's neural machine translation system: Bridging the gap
  between human and machine translation.
\newblock \emph{arXiv:1609.08144}.

\bibitem[{Xia et~al.(2018)Xia, Zhang, Yan, Chang, and Yu}]{xia-etal-2018-zero}
Congying Xia, Chenwei Zhang, Xiaohui Yan, Yi~Chang, and Philip Yu. 2018.
\newblock \href {https://doi.org/10.18653/v1/D18-1348} {Zero-shot user intent
  detection via capsule neural networks}.
\newblock In \emph{Proceedings of the 2018 Conference on Empirical Methods in
  Natural Language Processing}, pages 3090--3099, Brussels, Belgium.
  Association for Computational Linguistics.

\bibitem[{Xu and Sarikaya(2013{\natexlab{a}})}]{xu2013convolutional}
Puyang Xu and Ruhi Sarikaya. 2013{\natexlab{a}}.
\newblock Convolutional neural network based triangular {CRF} for joint intent
  detection and slot filling.
\newblock In \emph{2013 IEEE Workshop on Automatic Speech Recognition and
  Understanding}, pages 78--83. IEEE.

\bibitem[{Xu and Sarikaya(2013{\natexlab{b}})}]{xu2013exploiting}
Puyang Xu and Ruhi Sarikaya. 2013{\natexlab{b}}.
\newblock Exploiting shared information for multi-intent natural language
  sentence classification.
\newblock In \emph{Interspeech}, pages 3785--3789.

\bibitem[{Yang et~al.(2019)Yang, Dai, Yang, Carbonell, Salakhutdinov, and
  Le}]{yang2019xlnet}
Zhilin Yang, Zihang Dai, Yiming Yang, Jaime~G. Carbonell, Ruslan Salakhutdinov,
  and Quoc~V. Le. 2019.
\newblock \href
  {https://proceedings.neurips.cc/paper/2019/hash/dc6a7e655d7e5840e66733e9ee67cc69-Abstract.html}
  {Xlnet: Generalized autoregressive pretraining for language understanding}.
\newblock In \emph{Advances in Neural Information Processing Systems 32: Annual
  Conference on Neural Information Processing Systems 2019, NeurIPS 2019,
  December 8-14, 2019, Vancouver, BC, Canada}, pages 5754--5764.

\bibitem[{Zhang et~al.(2019)Zhang, Li, Du, Fan, and Yu}]{zhang-etal-2019-joint}
Chenwei Zhang, Yaliang Li, Nan Du, Wei Fan, and Philip Yu. 2019.
\newblock \href {https://doi.org/10.18653/v1/P19-1519} {Joint slot filling and
  intent detection via capsule neural networks}.
\newblock In \emph{Proceedings of the 57th Annual Meeting of the Association
  for Computational Linguistics}, pages 5259--5267, Florence, Italy.
  Association for Computational Linguistics.

\end{thebibliography}

\end{document}